\documentclass[submission,copyright,creativecommons]{eptcs}
\providecommand{\event}{ICLP 2020 - Women in LP} 
\usepackage{breakurl}             
\usepackage{underscore}           

\usepackage{mathptmx}

\usepackage[most]{tcolorbox}
\usepackage{makeidx}
\usepackage[english]{babel}
\usepackage{amsmath}
\usepackage{amsfonts}
\usepackage{amssymb}
\usepackage{graphicx}
\usepackage{xspace}
\usepackage{url}
\usepackage{algorithm}
\usepackage{algpseudocode}
\usepackage{booktabs}
\usepackage{multirow}
\usepackage[labelsep=quad,indention=10pt]{subfig}
\usepackage{bbding}
\usepackage{todonotes}
\usepackage{adjustbox}
\usepackage{pgfplots}
\usepackage{rotating}
\usepackage{wrapfig}

\newcommand{\sysfont}{\textit}
\newcommand{\lpopt}{\sysfont{lpopt}\xspace}

\newcommand{\clingo}{\sysfont{clingo}\xspace}

\newcommand{\dlv}{\sysfont{DLV}\xspace}
\newcommand{\idlv}{{{\small $\cal I$}-}\dlv}

\newcommand{\idlvd}{\idlv$_{always}$\xspace}
\newcommand{\idlvnd}{\idlv$_{never}$\xspace}
\newcommand{\idlvhd}{\idlv$_{deduct}$\xspace}
\newcommand{\idlvml}{\idlv$_{induct}$\xspace}

\newcommand{\derives}{\mbox{\,:\hspace{0.1em}\texttt{-}}\,\xspace}

\newcommand{\anonym}{\text{\underline{\hspace{0.2cm}}}}

\newcommand{\cit}[1]{~\cite{#1}}
\newcommand{\re}[1]{~\ref{#1}}

\newcommand{\commentsymbolright}{/*}
\newcommand{\commentsymbolleft}{*/}
\algrenewcommand\algorithmiccomment[1]{\hfill \commentsymbolright{} #1 \commentsymbolleft{}}


\newenvironment{dlvcode}
  {\vspace{-0.2cm}\begin{displaymath}\begin{array}{l}}
  {\end{array}\end{displaymath}\vspace{-0.3cm}}

\newcommand{\ndecomp}{``do-not-decomp"\xspace}
\newcommand{\decomp}{``decomp"\xspace}
\newcommand{\indifferent}{``indifferent"\xspace}

\newcommand{\join}[1]{\#sharedVars(#1)}

\newcommand{\arity}[1]{$arity(#1)$}
\newcommand{\body}[1]{B(#1)}
\newcommand{\He}[1]{H(#1)}

\newcommand{\len}[1]{|#1|}
\newcommand{\blen}[1]{\len{\body{#1}}}
\newcommand{\td}[1]{RD(#1)}

\newcommand{\FIlen}{$\len{Facts^I(P)}$}
\newcommand{\PIDB}{IDB(P)}

\title{A Machine Learning guided Rewriting Approach for ASP Logic Programs
\thanks{
This work has been partially supported by MIUR under project ``Declarative Reasoning over Streams'' (CUP H24I17000080001) -- PRIN 2017, by MISE under project ``S2BDW'' (F/050389/01-03/X32) -- ``Horizon2020'' PON I\&C2014-20, by Regione Calabria under project ``DLV LargeScale'' (CUP J28C17000220006) -- POR Calabria 2014-20.
}
}
\author{
Elena Mastria
\institute{Department of Mathematics and Computer Science\\
University of Calabria, Italy}
\email{elena.mastria@unical.it}
\and
Jessica Zangari
\institute{Department of Mathematics and Computer Science\\
University of Calabria, Italy}
\email{zangari@mat.unical.it}
\and
Simona Perri
\institute{Department of Mathematics and Computer Science\\
University of Calabria, Italy}
\email{perri@mat.unical.it}
\and
Francesco Calimeri
\institute{Department of Mathematics and Computer Science\\
University of Calabria, Italy}
\email{calimeri@mat.unical.it}
}
\def\titlerunning{A Machine Learning guided Rewriting Approach for ASP Logic Programs}
\def\authorrunning{E. Mastria, J. Zangari, S.Perri, F. Calimeri}

\begin{document}

\maketitle

\begin{abstract}

Answer Set Programming (ASP) is a declarative logic formalism that allows to encode computational problems via logic programs.
Despite the declarative nature of the formalism, some advanced expertise is required, in general, for designing an ASP encoding that can be efficiently evaluated by an actual ASP system.
A common way for trying to reduce the burden of manually tweaking an ASP program consists in automatically rewriting the input encoding according to suitable techniques, for producing alternative, yet semantically equivalent, ASP programs.
However, rewriting does not always grant benefits in terms of performance; hence, proper means are needed for predicting their effects with this respect.
In this paper we describe an approach based on Machine Learning (ML) to automatically decide
whether to rewrite.
In particular, given an ASP program and a set of input facts, our approach chooses whether and how to rewrite input rules based on a set of features measuring their structural properties and domain information.
To this end, a Multilayer Perceptrons model has then been trained to guide the ASP grounder \idlv on rewriting input rules.
We report and discuss the results of an experimental evaluation over a prototypical implementation.
\end{abstract}


\section{Introduction}\label{sec:intro}
Answer Set Programming (ASP)~\cite{DBLP:journals/cacm/BrewkaET11,DBLP:journals/ngc/GelfondL91} is a declarative programming paradigm proposed in the area of non-monotonic reasoning and logic programming.
With ASP, computational problems are encoded by logic programs whose intended models, called answer sets, correspond one-to-one to solutions of the original problem.
After several years of theoretical research, the scientific community reached a general consensus regarding the foundations of ASP computation, and a number of efficient evaluation methods and real systems are available today~\cite{perri-ijcai-2018-survey,DBLP:journals/ai/CalimeriGMR16}.
%
%
Typically, the same computational problem can be encoded by means of many different ASP programs which are semantically equivalent; however, real ASP systems may perform very differently when evaluating each one of them.

Indeed, structural properties of a logic program can make computation easier or harder; furthermore, specific aspects and features of the ASP system at hand might have significant impact on performance, as, for instance, adopted algorithms and optimizations.
As a result, some expert knowledge can be required in order to select the ``best'' encoding when performance is crucial; this, in a certain sense, conflicts with the declarative nature of ASP that, ideally, should free users from the burden of computational issues.
For this reason, ASP systems tend to be endowed with pre-processing means aiming at making performance less encoding-dependent; intuitively, this also fosters the usage of ASP in practice.

The idea of transforming logic programs has been explored in past literature, to different extents, such as verification, performance improvements, etc. (see e.g.,\cit{DBLP:journals/csur/PettorossiP96,DBLP:conf/tacas/AngelisFPP14} and related works);  %
in this paper, we focus on ASP and sketch a preliminary work on a Machine Learning (ML) strategy for automatically optimizing ASP encodings.
%
Such strategy relies on an adaptation of hypergraph tree decompositions techniques for rewriting rules and inductively estimates whether decomposition is convenient or not.
We devised and experimentally tested a prototypical implementation relying on the system \idlv~\cite{DBLP:journals/ia/CalimeriFPZ17}.
Experimental results show that the approach is promising; indeed, despite the embryonal nature of the system, performance are already comparable to the ones gained with well-assessed deductive heuristics.
%


\section{Tree Decompositions for Rewriting ASP Rules}\label{sec:TD}



An ASP rule $r$ can be represented as a \textit{hypergraph}\cit{DBLP:journals/tplp/BichlerMW16} $HG(r)$ and be decomposed according to a tree decomposition $TD(r)$ of $HG(r)$ into a set $RD(r)$ of new rules that are equivalent to the original one, yet typically shorter. Such technique is adopted in \lpopt\cit{DBLP:journals/tplp/BichlerMW16} to rewrite a program before it is fed to an ASP system.
In more details, a (undirected) hypergraph is a generalization of a (undirected) graph in which an edge can join two or more vertices. $HG(r)$ has a hyperedge for each literal $l$ in the body and in the head of $r$ containing all variables in $l$.
A tree decomposition of a hypergraph $HG(r)$ is a tuple ($TD(r)$, $\chi$), where $TD(r)$ = $(V(TD(r)), E(TD(r)))$ is a tree and $\chi :$ $V(TD(r)) \rightarrow$ $2^{V(HG(r))}$ is a function mapping a set of vertices $\chi(t) \subseteq V(HG(r))$ to each vertex $t$ of the decomposition tree $TD(r)$, such that for each $e \in E(HG(r))$ there is a node $t \in V(TD(r))$ such that $e \subseteq \chi(t)$, and for each $v \in  V(HG(r))$ the set $\{t \in V(TD(r)) | v \in \chi(t)\}$ is connected in $TD(r)$.
Intuitively, a tree decomposition $TD(r)$ of $HG(r)$ is a tree such that each vertex is associated to a \emph{bag}, i.e., a set of nodes of $HG(r)$, and such that each hyperedge of $HG(r)$ is covered by some bag, and for each node of $HG(r)$ all vertices of $TD(r)$ whose bag contains it induce a connected subtree of $TD(r)$.
%

A tree decomposition $TD(r)$ induces a set of rules that rewrites $r$, called {\em rule decomposition}, and denoted by $RD(r)$ containing a fresh rule for each vertex $v$ of $TD(r)$. Roughly, each body literal $l$ in $r$, such that the set of variables in $l$ is contained in $v$, is added to the body of the rule generated for $v$. Moreover, some rules may be generated to guarantee safety.
In general, more than one decomposition is possible for each rule.
The following example
illustrates these ideas.
Let us consider the rule:
\centerline{
$r_1: p(X,Y,Z,S)\derives s(S),\ a(X,Y,S-1),\ c(D,Y,Z),\
f(X,P,S-1),\ P >= D.$
}

\begin{figure}[!t]
    \centering
       \subfloat[$HG(r_1)$]{%
          \includegraphics[width=25.0mm]{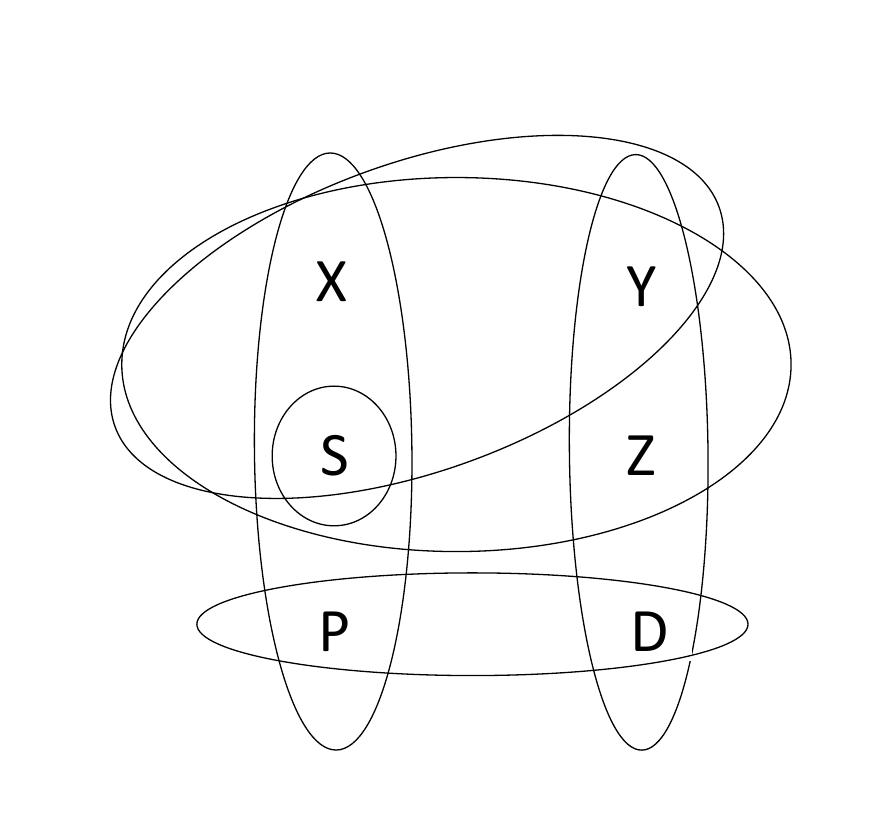}%
       }
       \subfloat[$TD_1(r_1)$]{%
          \includegraphics[width=25.0mm]{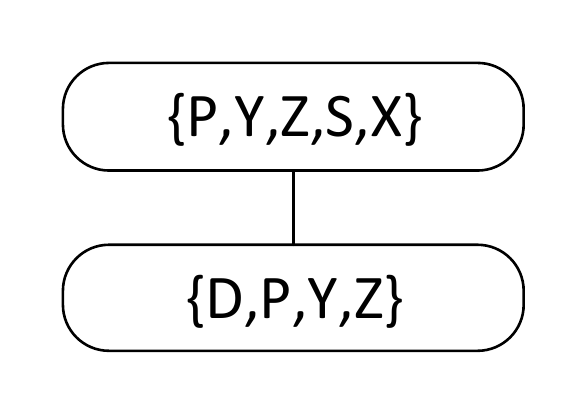}%
       }
       \subfloat[$TD_2(r_1)$]{%
          \includegraphics[width=25.0mm]{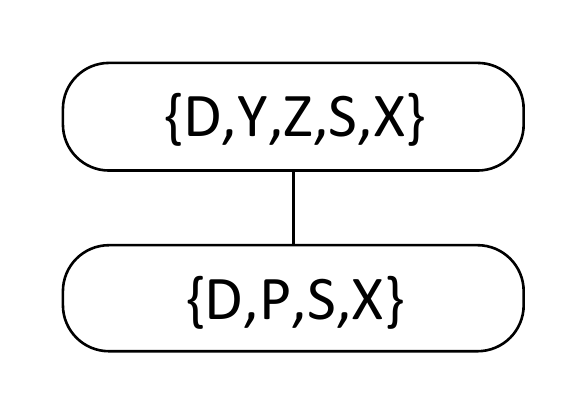}%
       }
       \caption{Decomposing a rule. \label{fig:nmi} } 
\vspace{-1\baselineskip}
\end{figure}
Figure\re{fig:nmi} depicts the conversion of rule $r_1$ into the hypergraph $HG(r_1)$ and two possible decompositions $TD_1(r_1)$ and $TD_2(r_1)$.
According to $TD_1(r_1)$, $r_1$ can be decomposed into the set of rules $RD_1(r_1)$:
%
\begin{dlvcode}
    r_2: p(X,Y,Z,S) \derives s(S),\ a(X,Y,S-1),\ f(X,P,S-1),\ fresh\_pred\_1(P,Y,Z).\\
    r_3: fresh\_pred\_1(P,Y,Z) \derives c(D,Y,Z),\ P>=D,\ fresh\_pred\_2(P).\\
    r_4: fresh\_pred\_2(P) \derives s(S),\ f(\anonym,P,S-1).
\end{dlvcode}

\noindent The rule $r_2$ has the same head of $r_1$ and in the body all literals covering the first node of $TD_1(r_1)$ with variables $\{P,Y,Z,S,X\}$; $r_3$ has in the head the fresh predicate $fresh\_pred\_1$ that links it to $r_2$. The body of $r_2$ contains the literals having as variables $\{D,P,Y,Z\}$ appearing in the other node of $TD_1(r_1)$. The rule $r_4$ ensures safety in $r_3$: $fresh\_pred\_2(P)$ is added in the body of $r_3$ and to the head of $r_4$, whose body has a set of literals coming from $r_1$ and covers $P$. Intuitively, a different rewriting could be obtained by differently handling safety: e.g., by adding the literals $s(S)$ and $f(\anonym,P,S-1)$ to the body of $r_3$ and avoiding to introduce $r_4$.
Similarly, according to $TD_2(r_1)$, $r_1$ can be rewritten into $RD_2(r_1)$:
%
\begin{dlvcode}
    r_5: p(X,Y,Z,S) \derives a(X,Y,S-1),c(D,Y,Z), fresh\_pred\_1(D,S,X).\\
    r_6: fresh\_pred\_1(D,S,X) \derives s(S),f(X,P,S-1),P>=D,fresh\_pred\_2(D).\\
    r_7: fresh\_pred\_2(D) \derives c(D,\anonym,\anonym).
\end{dlvcode}


\section{ML-guided Tree Decomposition Rewriting}
The commonly adopted approach for the evaluation of ASP programs relies on a grounding (or instantiation) module (\emph{grounder}) that generates a propositional theory semantically equivalent to the input program coupled with a subsequent module (\emph{solver}), that uses propositional techniques for generating answer sets.
There are monolithic systems integrating both computational stages such as \dlv~
\cite{DBLP:conf/lpnmr/AlvianoCDFLPRVZ17} and \clingo~\cite{DBLP:journals/tplp/GebserKKS19}, as well as systems performing only grounding or stand-alone solvers.

Among grounders, \idlv~\cite{DBLP:journals/ia/CalimeriFPZ17} employs a heuristic-guided tree decomposition algorithm~\cit{calimeri_perri_zangari_2019-TPLP} aiming at optimizing the instantiation process. Roughly, \idlv possibly decomposes input rules into multiple smaller ones according to the technique sketched in Section\re{sec:TD} on the basis of a formulas that estimates the cost of joining body literals. This deductive heuristic relies on internally computed grounding statistics such as the number of generated atoms or arguments's selectivities (cfr.~\cit{calimeri_perri_zangari_2019-TPLP}). The technique proved beneficial on both grounding and solving performance, permitting to mitigate the so-called grounding bottleneck issue and to actually instantiate programs that cannot be grounded otherwise.

In this paper, in contrast with the aforementioned proposal, we present an approach relying on an inductive heuristic. We still aim at properly deciding whether decomposing rules might improve grounding performance but via a ML heuristic that considers only ``static" information on the non-ground structure of the input program. In particular, this new heuristic is based on a predictive model purposefully designed and trained able to classify each input rule as: ``better to decompose", ``better not to decompose" or \indifferent (i.e., applying or not decompositions has almost the same effect on performance).

\subsection{Work-flow}

As anticipated, we opted for classification. We recall that in such a task input data consists of a set of records known as \textit{examples}; each record (or {\it example}) is a tuple of form $(X,y)$, where $X=\{x_1,\cdots,x_n\}$ for $n>0$ is a set of attributes and $y$ is a class label, namely, the so-called {\it target attribute}. Classification lies in learning a target function $f: X\mapsto y$ that maps a set of attributes $X$ to a class label $y$. In our context, for a rule $r$, $X$ is the features described in Section\re{subsec:features} computed on $r$, while $y$ can be either ``better to decompose", ``better not to decompose" or \indifferent.

We chose a set of features by focusing only on easily computable non-ground structural properties and domain information. The set of examples has been created by selecting all decomposable rules from a large set of widely spread ASP benchmarks.
We obtained an example from each selected rule by computing features on it and associating a class label. The association has been done by taking into account the \idlv grounding times when the considered rule is decomposed or not (see Section\re{sub:dataset}).
Once we achieved a consistent data set, an Artificial Neural Network (ANN) has been designed and trained to build the classifier. Eventually, we experimentally evaluated the quality of the resulting model.

\subsection{Feature Selection}\label{subsec:features}
In total, we devised $19$ features involving information about non-ground properties, tree decomposition structures and input facts.
For brevity, we herein focus on the $6$ features that showed a higher correlation with class labels. The features, reported below, are defined via the following notations.
Let $P$ be an ASP program, we indicate as $Facts^I(P)$ the set of input facts of $P$. We denote as $EDB(P)$ the set of all predicates in $P$ defined only by facts and as  $\PIDB$ the remaining ones.
Let $r$ be a rule, $\He{r}$ is the set of atoms in the {\em head} of $r$, as $\body{r}$ the set of literals in {\em body} of $r$ and as $\td{r}$ a possible tree decomposition of $r$.
We denote as $\join{r}$ the number of joins in $B(r)$
i.e. the number of times each pair of atoms in the rule shares the same variable.
Let $p$ be a predicate, we indicate as \arity{p} the arity of $p$.
\noindent

\begin{table}[!h]
\vspace{-0.5\baselineskip}
\centering
\begin{tabular}{c c c c c c}
	\FIlen &
	$\blen{r}$ &
	$\len{\td{r}}$ &
	$\frac{\sum\limits_{r_i\in \td{r}} |B(r_i)|}{|\td{r}|}$ &
	$\sum\limits_{r_i\in \td{r}} \join{r_i}$&
	$\frac{\sum\limits_{p_i \in \PIDB} arity(p_i)}{\len{\PIDB}}$
\end{tabular}
\vspace{-0.8\baselineskip}
\end{table}

Intuitively, for a rule $r$, from left to right, the $6$ features concern: $(i)$ the number of input facts, $(ii)$ the body length, $(iii)$ the number of rules in which $r$ can be decomposed, $(iv)$ the average length and $(v)$ the total number of joins in the rule decomposition, $(vi)$ the average arity of \textit{IDB} predicates.
\subsection{Data Set Creation}\label{sub:dataset}
We collected decomposable rules from benchmarks of the 3th, the 4th and the 6th official ASP competitions as well as from the grounding-intensive 2-QBF domain\cit{DBLP:journals/tplp/BichlerMW16}.
Since often rules cannot be decomposed as tree decomposition is applicable only on  the basis of intrinsic structural rule properties, we tried to enrich the set of examples. To this end, we performed a preprocessing step in which we applied to encodings techniques inspired by \emph{unfolding}\cit{seki1991unfold}. The aim was to obtain additional rules with longer bodies which are more likely to be decomposable.

At this point, we computed the features over the so collected decomposable rules and then, we properly associated a class label to each one.
Labels have been assigned by first generating for each considered rule an {\em example program} consisting of its non-decomposed version along with the set of required input facts, and then by measuring \idlv times when asked to instantiate the example program when decomposition is disabled and forcibly activated.
If the difference in instantiation times of two versions is lower than $10\%$, the features set is labelled as \indifferent; otherwise, the assigned label is either \decomp or \ndecomp, depending on which version led to lowest grounding time.

The resulting data set is formed by $3852$ examples: $3106$ have been labeled as \indifferent, $417$ as \ndecomp and the remaining $329$ as \decomp, with class distributions of $80,63\%$, $10,83\%$ and $8,54\%$, respectively.
We observe unbalanced distributions that, in general, could make learning about minority classes more tough and in turn, worsen the model quality. In Section\re{subsec:model} we describe our countermeasures to mitigate this issue.
\subsection{Model Design}\label{subsec:model}
To build the classifier, we adopted a \textit{Multilayer Perceptrons (MLPs) Neural Network}\cit{daniel2013principles}.
MLPs are commonly used for such tasks, as they often permit to reach a high accuracy by ``learning'' complex implicit  relationships within data.
In classifications, the learning process of a neural network is guided by a {\em loss function} that, in general, determines the quality of the model prediction.
More specifically, during the training phase, the internal network configuration undergoes through a series of transformations aiming at minimizing the loss function. Since it is computed at each training step, the loss value gives a clear indicator of how well the current configuration performs the task for which the network was designed; in our case, a multi-class classification.

Given that we wanted to maintain the natural data set configuration, we adopted a cost-sensitive learning method to deal with the imbalance issue.
Such approach, instead of modifying the distribution of training data, assigns different weights to classes in the loss function so that minority classes misclassifications are more penalized; this commonly used adjustment allows to deal with the imbalance directly into the learning algorithm itself.
In particular, we implemented as custom loss function the \emph{$\alpha$-balanced focal loss} for multi-classes classification\cit{FocalLoss}.
Experimentally, this focal loss variant has proved to be suitable for better classifying examples belonging to minority classes.
For the loss function minimization process, we used the adaptive learning rate optimization algorithm \textit{Adam} as an optimizer.
\subsection{Model Evaluation}

\begin{figure}[t]
\begin{center}
	\subfloat[Performance measures]{
	\begin{tiny}
	\begin{minipage}[c][]{
	   0.28\textwidth}
    	\centering
    	\vspace{5em}
    	\setlength{\tabcolsep}{0.4em}
		\begin{tabular}{|c||c|c|c|}
 		\hline
 		Class & Precision& Recall& F1-\textit{score}
 		\\ [0.5ex]
 		\hline\hline
 		\indifferent & $0.97$ & $0.96$ &$0.96$ 
 		\\
  		\ndecomp & $0.86$ & $0.83$ &$0.85$ 
  		\\
 		\decomp & $0.78$ & $0.88$ & $0.82$ 
 		\\
 		\hline
 		\multicolumn{4}{c}{Cumulative}\\
 		\hline
 		Avg method & Precision& Recall& F1-\textit{score}
 		\\ [0.5ex]
 		\hline\hline
 		macro & $0.87$ & $0.89$ &$0.88$ 
 		\\
 		weighted & $0.94$ & $0.94$ &$0.94$ 
 		\\
		\hline
		\end{tabular}
		\label{tab:val_measures}
	\end{minipage}
	\end{tiny}}
	\hfill
  \subfloat[Confusion Matrix]{
	\begin{minipage}[c][]{
	   0.28\textwidth}
	   \centering
	   \vspace{25pt}

	   \includegraphics[scale=0.30]{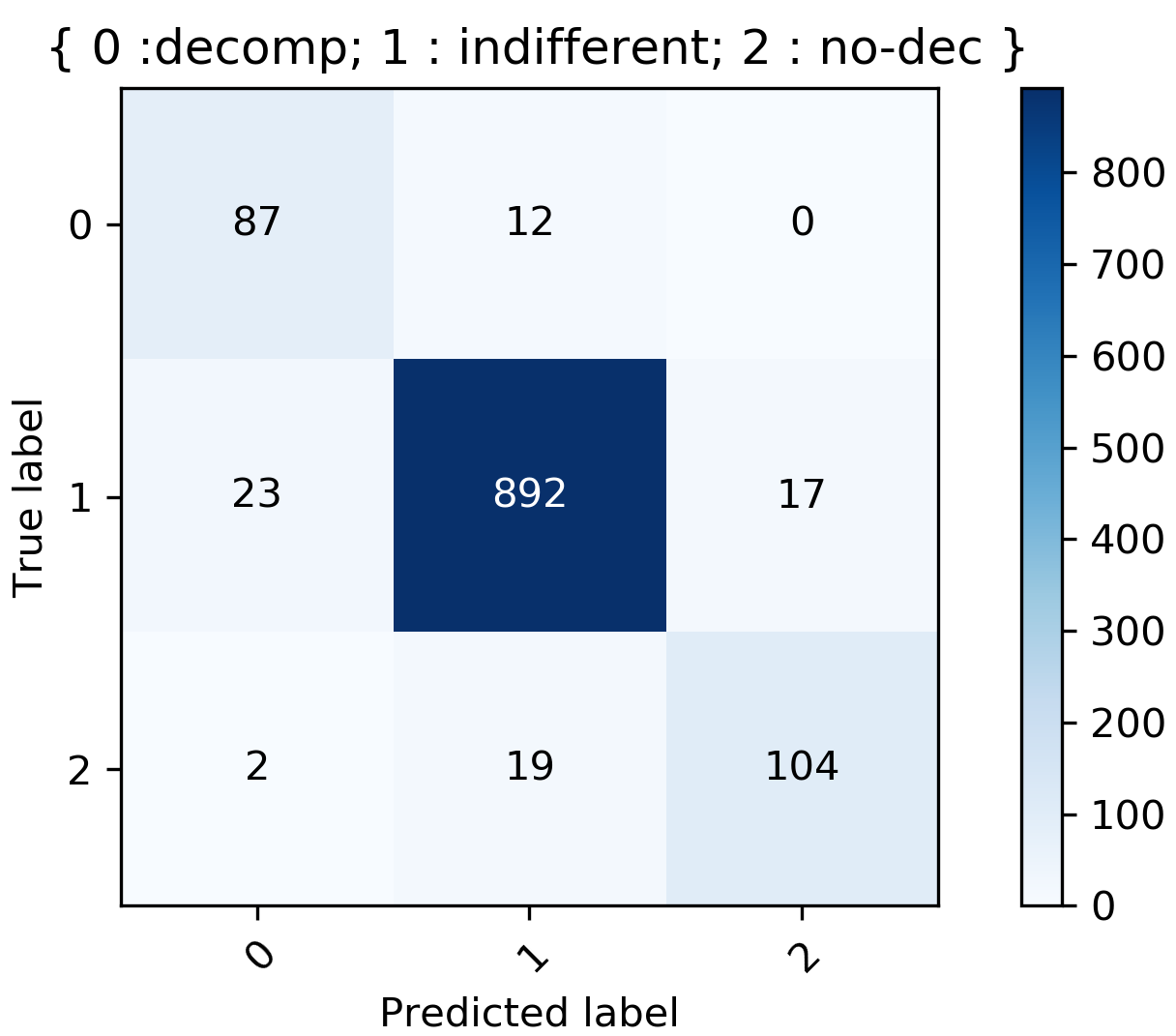}
	   	   \label{fig:cf_matrix}
	\end{minipage}}
 \hfill 	
  \subfloat[AUC-ROC curves]{
	\begin{minipage}[c][]{
	   0.28\textwidth}
	   \centering
	   \includegraphics[scale=0.30]{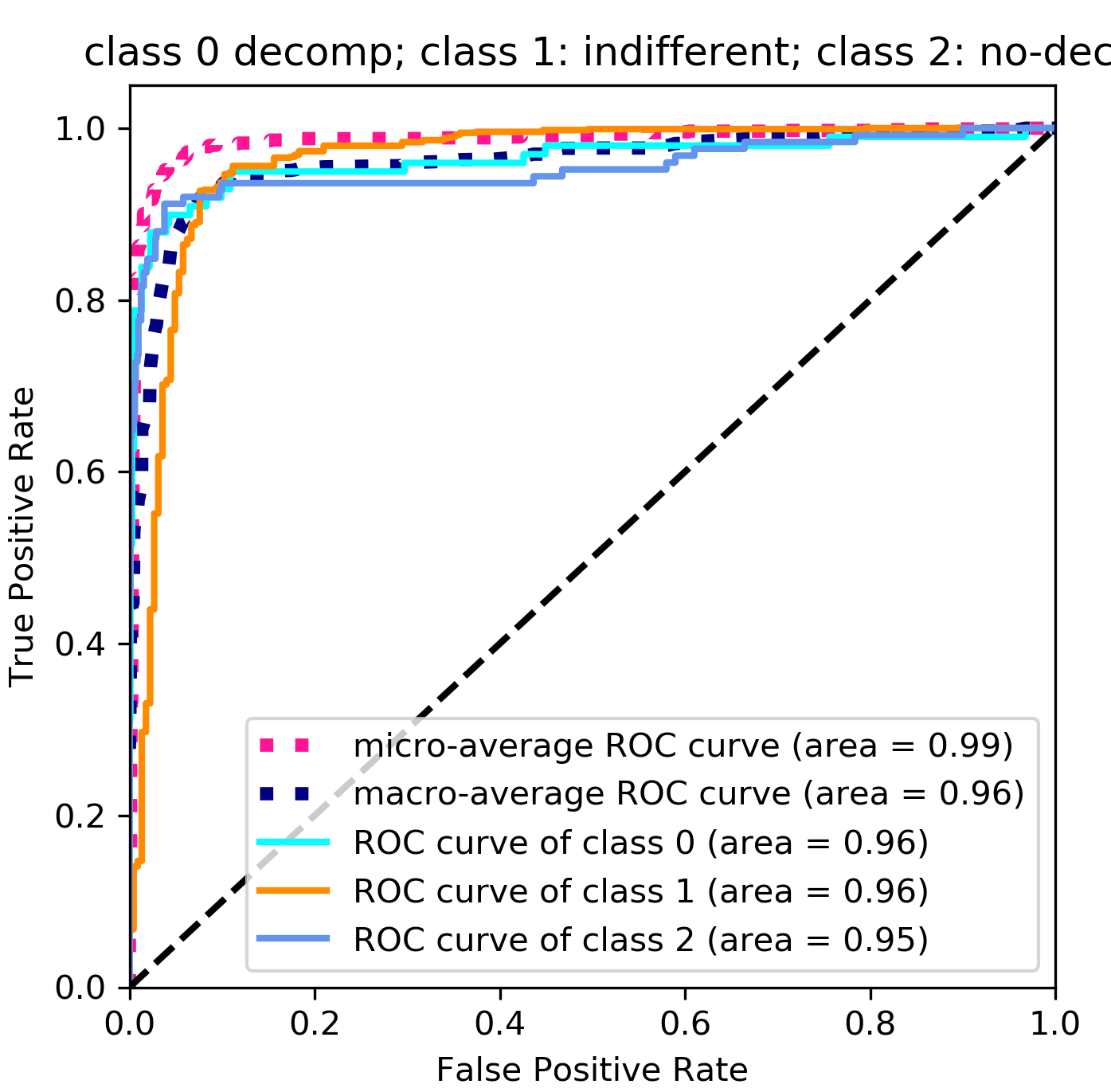}
	   \label{fig:roc_curve}
	\end{minipage}}
	\hfill
	\vspace{5pt}
	\caption{Model validation results.}
\label{fig:cfandroc}
\end{center}
\vspace{-1\baselineskip}
\end{figure}

The model has been trained over $300$ epochs. As convention, $70\%$ of the examples have been used as \textit{training set} and the remaining $30\%$ as \textit{test set}. In this splitting we carefully maintained the original class distributions.
Since we are dealing with an unbalanced data set, accuracy cannot be considered as an appropriate performance measure. In this metric, the impact of the classification errors of minority cases is reduced by the proper classification of majority cases. Thus, the quality of the model has been assessed by means of the \textit{F1-score} defined as $F1=2\;\times\;(Precision$ $ \times\; Recall)/{(Precision+Recall)}$ which provides us with more information about the effectiveness of the model on correctly predicting the instances belonging to minority classes \cit{MeasuresImbalancedDomains}. The $F1$ value is high when both $Precision$ and $Recall$ are high; the former indicates the proportion of cases classified as relevant that are actually relevant, while, the latter measures the proportion of relevant cases classified among all the relevant ones.

The Receiver Operator Characteristic ({\em ROC}) curve
examines the model capability of detecting {\em True Positives (TP)} instances and compares it with {\em False Positive (FP)} predictions.
The ROC curve plots \textit{TP rate} against \textit{FP rate} on the vertical axis and on the horizontal axis, respectively.
The larger the Area Under the Curve ({\em AUC}) is, the higher is the quality of the model (i.e., AUC=1.0).

Figure\re{tab:val_measures} reports \emph{precision}, \emph{recall} and \emph{F1} scores both class by class and aggregated using macro and weighted as average methods. Despite unbalance, the model achieves good performance also when dealing with minority classes. Figure\re{fig:cf_matrix} shows the \emph{confusion matrix} summarizing distributions of model predictions.
The \emph{AUC - ROC} plot in Figure\re{fig:roc_curve} evidences the model capability of distinguishing among classes: an AUC very close to $1.0$ suggests that in most cases the model correctly identifies when it is convenient to decompose a rule or not.

\section{Experimental Evaluation and Conclusions}

\begin{table}[t]
\begin{footnotesize}
  \caption{4th Competition - number of grounded instances and average grounding times in seconds.}
  \label{tab:results}
  \begin{minipage}{\textwidth}
  \centering
      	\setlength{\tabcolsep}{0.4em}
      	\resizebox{33em}{!}{%
    \begin{tabular}{ll|rr|rr|rr|rr}
      \hline\hline
      \multicolumn{2}{c|}{Problem} & \multicolumn{2}{c|}{\idlvnd}& \multicolumn{2}{c|}{\idlvd} & \multicolumn{2}{c|}{\idlvhd} & \multicolumn{2}{c}{\idlvml}\\
      Name & \# instances & \# & Time & \# & Time & \# & Time & \# & Time\\
      \hline
		Permutation Pattern Matching & 30 & 28 & 58.69 & 30 & 63.90 & 30 & 64.83 & 30 & 63.89\\
		Valves Location & 30 & 30 & 4.11 & 30 & 4.09 & 30 & 4.10 & 30 & 4.11\\
		Connected Still Life & 10 & 10 & 0.10 & 10 & 0.10 & 10 & 0.10 & 10 & 0.10\\
		Graceful Graphs & 30 & 30 & 0.38 & 30 & 0.38 & 30 & 0.39 & 30 & 0.38\\
		Bottle Filling Problem & 30 & 30 & 4.07 & 30 & 4.31 & 30 & 4.33 & 30 & 4.33\\
		\textbf{Nomystery} & \textbf{30} & \textbf{30} & \textbf{34.86} & \textbf{30} & \textbf{18.92} & \textbf{30} & \textbf{35.66} & \textbf{30} & \textbf{18.84}\\
		Sokoban & 30 & 30 & 2.68 & 30 & 2.76 & 30 & 2.77 & 30 & 2.85\\
		Ricochet Robots & 30 & 30 & 0.27 & 30 & 0.31 & 30 & 0.31 & 30 & 0.31\\
		Crossing Minimization & 30 & 30 & 0.10 & 30 & 0.10 & 30 & 0.10 & 30 & 0.10\\
		Reachability & 30 & 30 & 102,46 & 30 & 101,65 & 30 & 102,72 & 30 & 102,72 \\
		Strategic Companies & 30 & 30 & 0,21 & 30 & 0,22 & 30 & 0,21 & 30 & 0,41 \\
		Solitaire & 27 & 27 & 0.13 & 27 & 0.19 & 27 & 0.20 & 27 & 0.20\\
		\textbf{Weighted-Sequence Problem} & \textbf{30} & \textbf{30} & \textbf{2.83 }& \textbf{30} & \textbf{9.50} & \textbf{30} & \textbf{2.90} & \textbf{30} & \textbf{11.16}\\
		Stable Marriage & 30 & 30 & 27.72 & 30 & 2.54 & 30 & 2.54 & 30 & 2.47\\
		Incremental Scheduling & 30 & 12 & 295.62 & 21 & 219.97 & 21 & 222.00 & 21 & 214.59\\
		Qualitative Spatial Reasoning & 30 & 30 & 2.84 & 30 & 2.84 & 30 & 2.83 & 30 & 2.85\\
		Chemical Classification & 30 & 30 & 88.49 & 30 & 88.50 & 30 & 88.67 & 30 & 87.28\\
		Abstract Dialectical Frameworks & 30 & 30 & 0.13 & 30 & 0.13 & 30 & 0.13 & 30 & 0.13\\
		Visit-all & 30 & 30 & 0.13 & 30 & 0.13 & 30 & 0.13 & 30 & 0.14\\
		Complex Optimization & 29 & 29 & 34.89 & 29 & 34.23 & 29 & 35.15 & 29 & 35.51\\
		Knight Tour with Holes & 30 & 20 & 177.76 & 20 & 173.03 & 20 & 174.90 & 20 & 180.98\\
		Maximal Clique & 30 & 30 & 0.32 & 30 & 0.32 & 30 & 0.33 & 30 & 0.31\\
		\textbf{Labyrinth} & \textbf{30} & \textbf{30} & \textbf{1.47} & \textbf{30} & \textbf{1.39} & \textbf{30} & \textbf{1.48} & \textbf{30} & \textbf{0.71}\\
		Minimal Diagnosis & 30 & 30 & 2.54 & 30 & 2.22 & 30 & 2.57 & 30 & 2.90\\
		Hanoi Tower & 30 & 30 & 0.22 & 30 & 0.23 & 30 & 0.23 & 30 & 0.23\\
		Graph Colouring & 30 & 30 & 0.10 & 30 & 0.10 & 30 & 0.10 & 30 & 0.10\\
		\hline
		Total & 696 & 666 & 22.46 & 677 & 22.39 & 677 & 23.07 & 677 & 22.47\\
      \hline\hline
    \end{tabular}}
    \vspace{-1\baselineskip}
  \end{minipage}
\end{footnotesize}
\end{table}

In this section, we analyze the impact of the proposed inductive heuristic on \idlv performance. Four versions of \idlv have been compared:
$(i) $ \idlvnd with decomposition disabled;
$(ii)$ \idlvd with decomposition always enabled;
$(iii)$ \idlvhd with decomposition applied according to the internal deductive heuristic;
$(iv)$ \idlvml with decomposition guided by the inductive model.
The latter version has been externally implemented thanks to the capability of \idlv to customize its grounding process via annotations\cit{DBLP:journals/ia/CalimeriFPZ17}. More in detail, the model communicates with \idlv via an external module which, given an encoding, for each rule $r$: first, invokes the model to determine whether $r$ has to be decomposed, and then accordingly, annotates $r$ as to decompose or not.
Eventually, these annotated encodings are fed to \idlv.
%
We report in Table\re{tab:results} results on the 4th competition. For each version, the table details the number of grounded instances and the average instantiation times per problem.
Experiments have been performed on a NUMA machine equipped with two 2.8 GHz AMD Opteron 6320 and 128 GiB of main memory, running Linux Ubuntu 14.04.4; memory limit has been set to 15 GiB and time limit to 600 seconds per instance.

In general, the proposed method behaves consistently with the well-established deductive method used in \idlv. On the one side, we observe cases such as \emph{Nomystery} in which the inductive heuristic identifies benefits of applying decompositions: \idlvhd rewrites the input encoding in a way similar to \idlvnd, while the \idlvml rewriting is comparable to that performed by \idlvd. An improvement is also gained in problem \emph{Labyrinth}: \idlvml is faster than the other versions. On the other side, there is the case of \emph{Weighted-Sequence Problem} in which the proposed method causes a significant worsening because it does not recognize some decompositions as convenient. In the remaining problems, \idlvml performance is in line with others.


%

In summary, one can note that, despite the embryonal nature of the work, performance are already comparable to the ones obtained with well-assessed methods.
Further studies will tell whether the inductive approach is actually effective for improving performance of ASP grounders.
In this respect, we plan to enrich the set of classification features, experimenting with other classification algorithms and considering a larger set of benchmarks for both training and testing.
Moreover, we plan to consider further rewriting techniques besides tree decomposition, and to extend our implementation with the capability of foreseeing effects of each single rewriting and/or combinations thereof.

\bibliographystyle{eptcs}
\bibliography{./references-cleaned}
\end{document}